\documentclass{article}

\usepackage{microtype}
\usepackage{graphicx}
\usepackage{booktabs} 

\usepackage{hyperref}

\usepackage[accepted]{icml2021}

\icmltitlerunning{Scalable pragmatic communication via self-supervision}

\usepackage{amsmath,amssymb}
\usepackage{bbm}
\usepackage{cleveref}
\usepackage{graphicx}
\graphicspath{{./figures/}}
\usepackage{floatrow}
\usepackage{subfig} 
\usepackage{helvet}
\usepackage{adjustbox}
\usepackage{tikz}
\newcommand{\colorcontext}[3]{
  \begin{tikzpicture}
    \draw[fill=#2] (-0.3,-0.3) rectangle (0.3,0.3);
    \node at (0, 0) {\color{white}{}};
    \draw[fill=#3] (0.3,-0.3) rectangle (0.9,0.3);
    \node at (0.6, 0) {\color{white}{}};
    \draw[fill=#1,ultra thick] (-0.9,-0.3) rectangle (-0.3,0.3); 
    \node at (-0.6, 0) {\color{white}{$m^*$}};
  \end{tikzpicture}
}
\newcommand{\HTMLcontext}[3]{
  \definecolor{color1}{HTML}{#1}
  \definecolor{color2}{HTML}{#2}
  \definecolor{color3}{HTML}{#3}
  \colorcontext{color1}{color2}{color3}
}

\usepackage{placeins}

\begin{document}

\twocolumn[
\icmltitle{Scalable pragmatic communication via self-supervision}



\icmlsetsymbol{equal}{*}

\begin{icmlauthorlist}
\icmlauthor{Jennifer Hu}{bcs}
\icmlauthor{Roger Levy}{bcs}
\icmlauthor{Noga Zaslavsky}{bcs,cbmm}
\end{icmlauthorlist}

\icmlaffiliation{bcs}{Department of Brain and Cognitive Sciences, Massachusetts Institute of Technology}
\icmlaffiliation{cbmm}{Center for Brains Minds and Machines, Massachusetts Institute of Technology}

\icmlcorrespondingauthor{Jennifer Hu}{jennhu@mit.edu}

\icmlkeywords{Pragmatics, Natural language processing, Emergent communication, Self-supervised learning, Machine learning, Rational Speech Act, Cognitive Science, Psycholinguistics, ICML}

\vskip 0.3in
]



\printAffiliationsAndNotice{}  

\begin{abstract}
Models of context-sensitive communication often use the Rational Speech Act framework \citep[RSA;][]{frank:goodman:2012}, which formulates listeners and speakers in a cooperative reasoning process. However, the standard RSA formulation can only be applied to small domains, and large-scale applications have relied on imitating human behavior. Here, we propose a new approach to scalable pragmatics, building upon recent theoretical results \citep{Zaslavsky:et-al:2020} that characterize pragmatic reasoning in terms of general information-theoretic principles. 
Specifically, we propose an architecture and learning process in which agents acquire pragmatic policies via self-supervision instead of imitating human data. This work suggests a new principled approach for equipping artificial agents with pragmatic skills via self-supervision, which is grounded both in pragmatic theory and in information theory. 
\end{abstract}

\section{Introduction}

When humans communicate, we do not simply transfer linguistic codes, but also integrate contextual knowledge and social expectations \citep{sperber:wilson:1986}. One prominent model of these \emph{pragmatic} phenomena is the Rational Speech Act framework \citep[RSA;][]{frank:goodman:2012}, which formulates speakers and listeners as rational agents in a cooperative signaling game.
While RSA has met broad empirical support in behavioral studies \citep[e.g.,][]{goodman:frank:2016},
applying RSA beyond small domains raises computational challenges. 
Large-scale applications of RSA have typically relied on training models to imitate human behaviors using contextually grounded datasets, such as utterances from humans describing a target image among a set of displayed images \citep[e.g.,][]{andreas:klein:2016,monroe-etal-2017-colors,mcdowell-goodman-2019-learning}. However, collecting such data can be difficult at scale, and requires the pre-determination of the types of contexts and domains in which artificial agents can reason pragmatically. How, then, might pragmatic reasoning skills be acquired in a theoretically principled and data-efficient manner? 

\begin{figure}[t]
    \centering
    \includegraphics[width=0.72\linewidth]{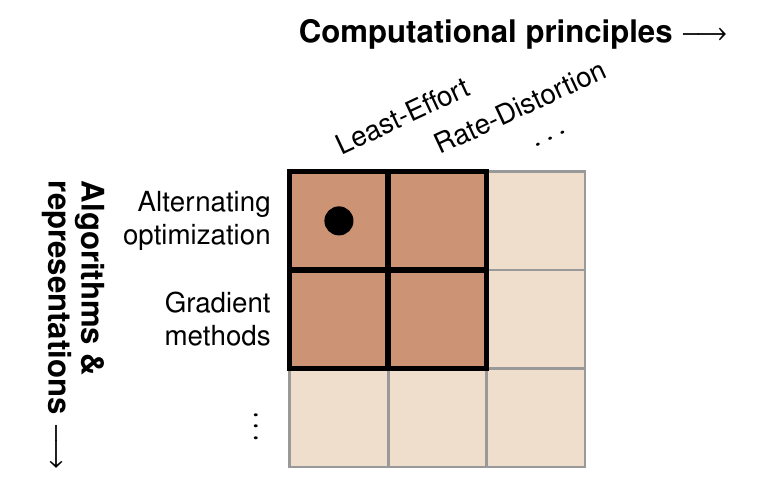}
    \caption{The RSA framework (black circle) is one instance in a more general model class, varying along two axes: computational principles, and algorithms and representations.}
    \label{fig:hypotheses}
\end{figure}

We address this question by leveraging recent findings that characterize pragmatic reasoning in terms of general information-theoretic optimization principles. \citet{Zaslavsky:et-al:2020} show that RSA implements alternating maximization of an objective function that may be viewed as a type of least-effort (LE) optimization principle. Furthermore, they show how to ground RSA in Rate--Distortion (RD) theory \citep{shannon_mathematical_1948,Shannon1959}, providing another candidate objective function. 
While RSA has traditionally been considered a computational-level account of human behavior in \citeauthor{marr_vision_1982}'s (\citeyear{marr_vision_1982}) sense, 
these findings suggest that RSA is only one instantiation of a more general model class, varying along two axes (\Cref{fig:hypotheses}): \emph{computational principles}, defined by objective functions such as LE and RD; and \emph{algorithms and representations} for optimizing these functions, such as alternating maximization (AM) or gradient descent (GD). 
AM, which is used in RSA, is intractable in large domains and does not generalize across domains. GD, on the other hand, is scalable and can potentially enable generalization across domains. However, to our knowledge, gradient-based methods have not previously been explored as an algorithmic implementation of pragmatic reasoning.

Based on these insights, we propose a novel approach to scalable pragmatics, in which agents acquire pragmatic policies via self-supervision. The key idea is to first learn a lexicon without contextual supervision, and then learn contextual pragmatic policies by optimizing a pragmatic objective function via GD, without any human supervision.
Using color reference games, we show that our method improves upon a context-agnostic model and achieves comparable performance to models trained with pragmatic supervision. 

\section{Background} \label{sec:background}

The Rational Speech Act framework \citep[RSA;][]{frank:goodman:2012} 
relates speakers and listeners as recursively reasoning about each other. 
Let $C$ be a referential context, which we take to be a subset of the full space of meanings, and $P(m)$ a prior distribution over meanings. Let $\mathcal{U}$ denote the set of possible utterances. The initial conditions of an RSA model are characterized by a lexicon $\mathcal{L}$, which takes an utterance $u \in \mathcal{U}$ and meaning $m \in C$ and outputs a value in $[0,1]$ representing the extent to which $u$ describes $m$. 
RSA models are typically grounded in a literal listener at recursion depth $t=0$, who defines a speaker-agnostic inference distribution over meanings given utterances:
\begin{align}
    l_0(m|u;C) &\propto \mathcal{L}(u,m) P(m) \,. \label{eq:am-rsa-l0}
\end{align}
The pragmatic speaker at $t\geq1$ defines a production distribution over utterances given an intended meaning by soft-maximizing a utility function, consisting of the surprisal of $l_{t-1}$ and the cost $\kappa(u)$ of producing $u$. That is,
\begin{align}
    s_t(u|m;C) &\propto \exp(\alpha(\log l_{t-1}(m|u;C) - \kappa(u)))\,. \label{eq:am-rsa-st}
\end{align}
The ``rationality'' parameter $\alpha \geq 0$ reflects the degree to which the speaker seeks to maximize utility.
Finally, the pragmatic listener at $t\geq1$ is Bayesian with respect to the pragmatic speaker at iteration $t$,
\begin{align}
    l_t(m|u;C) &\propto s_t(u|m;C) P(m)\,. \label{eq:am-rsa-lt}
\end{align}
\citet{Zaslavsky:et-al:2020} show that RSA implements an alternating maximization algorithm for optimizing
\begin{align}
    \mathcal{G}_{\alpha}[s,l] &= H_{s}(U|M) - \alpha \mathbb{E}_{s}\left[\log l(M|U)-\kappa(U)\right]\,. \label{eq:rsa-objective}
\end{align}
Because maximizing $H_s(U|M)$ amounts to minimizing the deviation from a random production distribution, maximizing $\mathcal{G}_\alpha$ can be viewed as a type of least-effort principle \citep{zipf_human_1949}, which we will refer to as LE-RSA.

By viewing speakers and listeners as probabilistic encoders and decoders under Rate--Distortion (RD) theory, \citeauthor{Zaslavsky:et-al:2020} argue that the optimal speaker and listener should optimize a tradeoff between maximizing utility and minimizing the mutual information between speaker utterances and meanings; i.e., minimizing $I(M;U)$ instead of maximizing $H(U|M)$. For reasons of space, we focus on LE-RSA here, but note that our models could easily be modified to optimize the RD-RSA objective.

\section{Color reference dataset} \label{sec:data}

We use \citeauthor{monroe-etal-2017-colors}'s (\citeyear{monroe-etal-2017-colors}) corpus of color reference games. 
On each round, pairs of human players saw a context of three colored squares in randomized order. The speaker's task was to describe a privately assigned target among the three colors, and the listener's task was to click on the inferred target based on the description. Both players could send messages through the chatbox at any time. Contexts were generated according to three conditions based on the perceptual distance between colors: 
\emph{far}, where all colors were dissimilar; \emph{split}, where exactly one distractor was similar to the target; and \emph{close}, where all colors were similar.

Exact AM is intractable in this domain. To compare AM and GD in a controlled manner, we simplified the corpus by removing listener messages, rounds with multiple speaker utterances, and casing and punctuation. We then collapsed spelling variants (e.g., ``gray''/ ``grey'') and only kept game rounds with the 100 most frequent utterances. 
The resulting data were split into 80\% train (18,288  rounds), 10\% validation, and 10\% test. See \Cref{tab:contexts} for examples.

\begin{table}[h]
    \centering
    \footnotesize
    \setlength{\tabcolsep}{1.5pt}
    \begin{tabular}{rcccccc} \toprule
        & Context & Human & Base & SSL-AM & SSL-GD & SL \\ \midrule
        $C_1$&\adjustbox{valign=c}{\trimbox{0.35cm 0cm 0cm -0.3em}{ \HTMLcontext{876e91}{884db3}{aa8455}}} & gray & purple & blue & gray & purple \\\rule{0pt}{0.6cm}
        $C_2$& \adjustbox{valign=c}{\trimbox{0.3cm -0.3em 0cm 0cm}{\HTMLcontext{86708f}{867c79}{7f817e}}} & purple & purple & red & purple & purple \\
        \bottomrule
    \end{tabular}
    \caption{Example human and model speaker utterances. The target color $m^*$ is nearly identical in both contexts, showing that human descriptions are context-sensitive.} 
    \label{tab:contexts} 
\end{table}

\section{Models}

Each game round is a tuple $(C, m^*, u, \hat{m}^*)$, where the context $C$ is an unordered triple of colors $\{m_1,m_2,m_3\}$; $m^* \in C$ is the target color; $u$ is the human speaker utterance; and $\hat{m}^*$ is the human listener selection. 
We represent all colors in CIELUV, a 3-dimensional perceptual color space designed for color displays. 
In what follows, we present our self-supervised learning (SSL) approach to pragmatics, as as well as a supervised learning approach that has previously been considered in this domain.

\subsection{Lexicon learning} \label{sec:colors-lex} 

As in the standard RSA formulation, our SSL approach assumes that agents have access to a shared lexicon, which reflects context-agnostic literal semantics. 
We capture this by learning a function $\mathcal{L}$, parameterized by $\theta$, which maps an utterance $u$ and color $m$ to a value in $(0,1)$, representing the extent to which $u$ applies to $m$. The utterance is first passed through an embedding layer,\footnote{Embeddings were initialized with off-the-shelf distilled RoBERTa~\citep{reimers-2019-sentence-bert} embeddings obtained from inputting ``This color is [utterance].'', projected to 50 dimensions using PCA, and fine-tuned during training.} and then the embedding and color are mapped by a single-layer feedforward neural network to a real-valued score, which is squashed by a sigmoid layer. 
This component is similar to \citeauthor{mcdowell-goodman-2019-learning}'s (\citeyear{mcdowell-goodman-2019-learning}) formulation of the lexicon. However, the key difference is that here we train the lexicon without supervision from human pragmatic behavior. Instead, our lexicon is trained by maximizing the likelihood of a \emph{decontextualized} version of the training set, which only provides isolated target colors and utterances without color contexts. The goal of this process is to learn semantics without contextual supervision, but we note that this is an approximation, as human participants did see contexts during gameplay. 

\subsection{Base agents} \label{sec:base}

After learning the lexicon, we define the RSA base listener and speaker as follows. Given a color context $C$, 
\begin{align}
    l_0(m|u,C;\theta) &\propto \mathcal{L}(u,m;\theta)P(m) \label{eq:colors-l-base} \\
    s_0(u|m,C;\theta) &\propto \mathcal{L}(u,m;\theta) \exp(-\kappa(u)) \;.
\end{align}
We take $P(m)$ to be uniform.
To capture utterance frequency effects, we take $\kappa(u) = -\log P(u)$, where $P(u)$ is the $n$-gram probability of $u$ in the Google Books corpus. 

\subsection{Self-supervised learning (SSL) pragmatic agents}

\paragraph{AM model.}
In RSA, the literal base agents are pragmatically enriched by iteratively updating \Cref{eq:am-rsa-st,eq:am-rsa-lt}. We argue that this can be viewed as a form of self-supervision, because this process implements AM for optimizing an agent-intrinsic objective (LE-RSA, \Cref{fig:hypotheses}). 
In this sense, the common approach of training the base agents with full (contextual) supervision and afterwards enriching them with AM steps~\citep[e.g.,][]{monroe-etal-2017-colors} can be viewed as partially self-supervised. For our SSL-AM agents we use $\alpha=1.17$ and $t=1$ AM iteration, which were tuned to maximize $l_t$ likelihood.

\paragraph{GD model.} The computations performed by the AM agents are independent across contexts and generally do not scale to large domains. We therefore propose an SSL-GD model, in which agents are parameterized by an architecture that 
may better generalize and scale to complex settings. In this model, the base agents are pragmatically enriched by taking a few GD steps, instead of AM steps, with respect to the same objective function.  
The listener model, parameterized by $\varphi$, has two trainable components: an utterance encoder $f_1: \mathbb{R}^3\to\mathbb{R}^3$ 
and a color context encoder $f_2: \mathbb{R}^{3|\mathcal{U}|} \to \mathbb{R}^3$. 
The speaker, parameterized by $\psi$, has two analogous components: a target color encoder $g_1: \mathbb{R}^{|\mathcal{U}|} \to \mathbb{R}^{|\mathcal{U}|}$ and a color context encoder $g_2:\mathbb{R}^{3|\mathcal{U}|} \to \mathbb{R}^{|\mathcal{U}|}$. $f_1$, $f_2$, $g_1$, and $g_2$ are parameterized by single-layer feed-forward neural networks. Listener and speaker distributions are read out by soft-maximizing the difference between the encodings of the input and context,
\begin{align}
    l^{GD}(m|u,C;\varphi) \propto \exp(&f_1(\mathcal{L}_u;\varphi)_m - f_2(\mathcal{L}_C;\varphi)_m \nonumber \\
    &+ \log\mathcal{L}(u,m;\theta)) \label{eq:colors-l-prag} \\
    s^{GD}(u|m,C;\psi) \propto \exp(&g_1(\mathcal{L}_m;\psi)_u - g_2(\mathcal{L}_C;\psi)^{T}_u \nonumber \\
    &+ \log\mathcal{L}(u,m;\theta)), \label{eq:colors-s-prag} 
\end{align}
where the encoders receive input representations of utterances, colors, and contexts as given by the lexicon $\mathcal{L}$, 
\begin{align}
    \mathcal{L}_u &= \langle \mathcal{L}(u,m_1;\theta), \mathcal{L}(u,m_2;\theta), \mathcal{L}(u,m_3;\theta) \rangle  \label{eq:lex-u}\\
    \mathcal{L}_m &= \langle \mathcal{L}(u_1,m;\theta), \mathcal{L}(u_2,m;\theta), \dots, \mathcal{L}(u_{|\mathcal{U}|},m;\theta) \rangle  \label{eq:lex-c}\\
    \mathcal{L}_C &= \langle \mathcal{L}_{m_1}, \mathcal{L}_{m_2}, \mathcal{L}_{m_3} \rangle. \label{eq:lex-context}
\end{align} 
To further avoid semantic drift, we add $\log\mathcal{L}(u,m;\theta)$ to the argument of the exponent, and the lexicon parameters $\theta$ are kept frozen. 
We initialize $\varphi$ and $\psi$ with values sampled uniformly from $(-0.01,0.01)$ and bias terms with 0. 

To derive model predictions for each context, $\varphi$ and $\psi$ are used to compute the pragmatic listener and speaker distributions, and then updated via GD to maximize the LE-RSA objective function~(\ref{eq:rsa-objective}). Since $\alpha$ parameterizes this function, we use the same value of $\alpha$ as in the AM model, and tune the remaining algorithm-specific hyperparameters on the validation set (9 GD steps, learning rate $\eta=0.357$).
$\varphi$ and $\psi$ are re-initialized for each context, in analogy to reasoning in the AM model. Crucially, our agents do not have access to human data during this phase. We note that this setup, in contrast to the AM model, also enables continual adaptation of the pragmatic parameters, which may facilitate knowledge transfer across contexts and domains. 

\begin{figure*}[t]
    \centering
    \raisebox{1cm}{\includegraphics[width=0.1\linewidth]{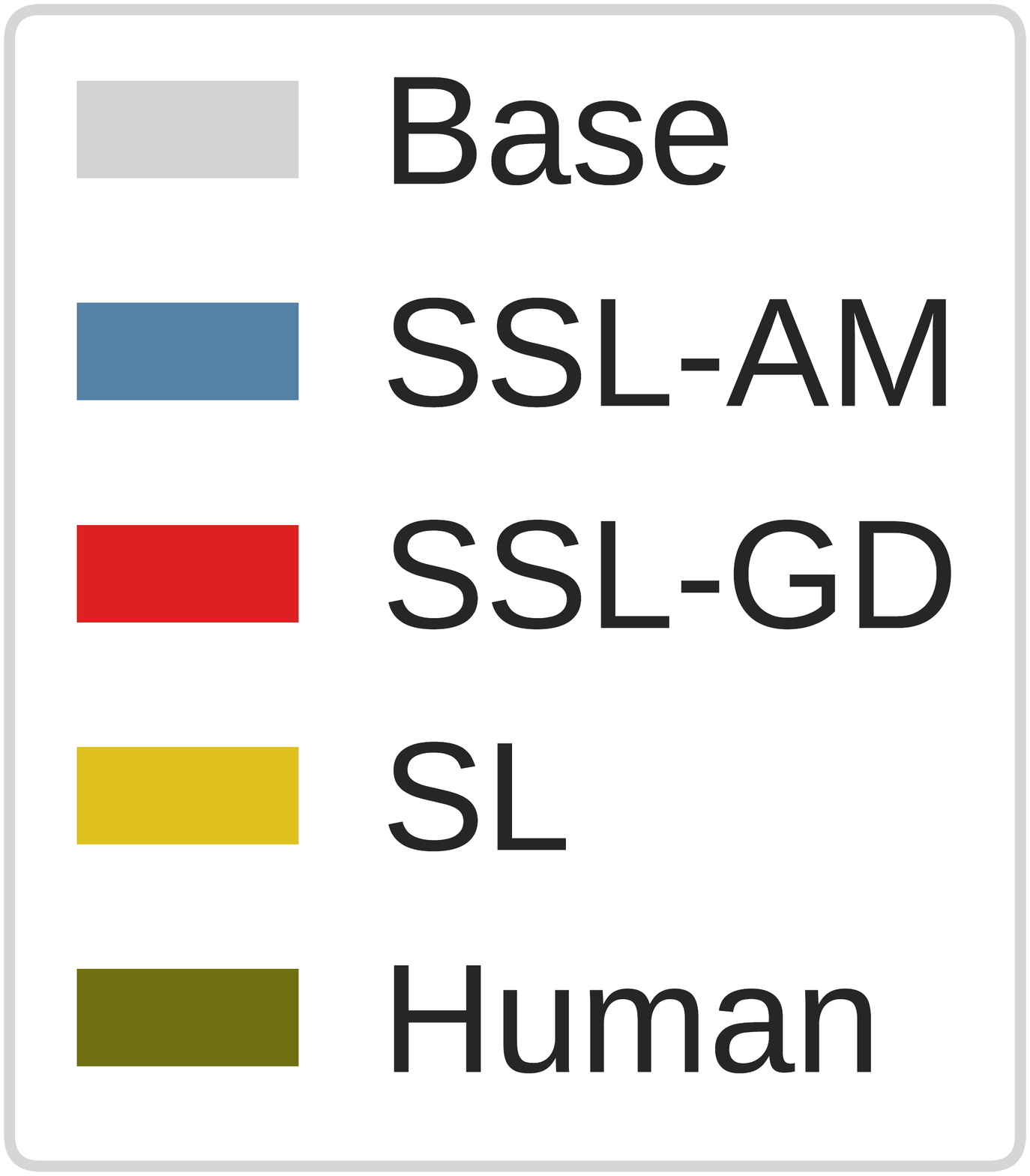}}
    \subfloat[]{\raisebox{0cm}{\includegraphics[width=0.33\linewidth]{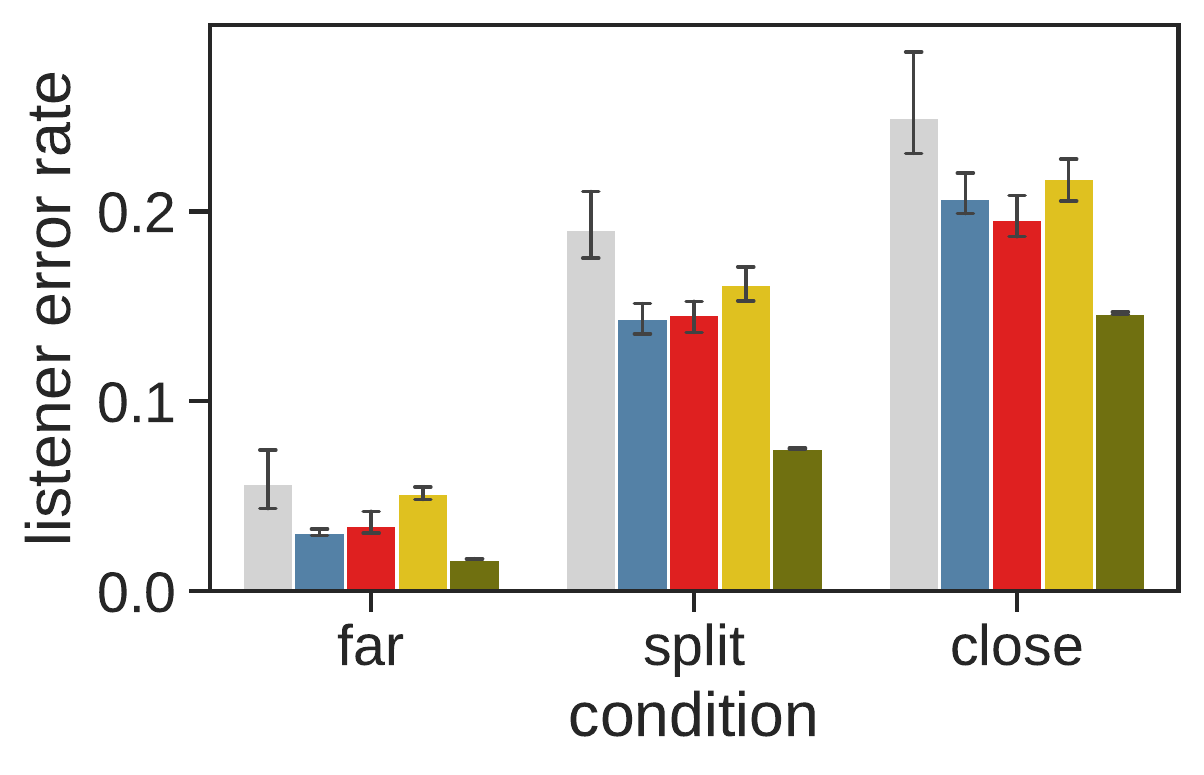}\label{fig:l-acc}}}~
    \subfloat[]{\raisebox{0cm}{\includegraphics[width=0.55\linewidth]{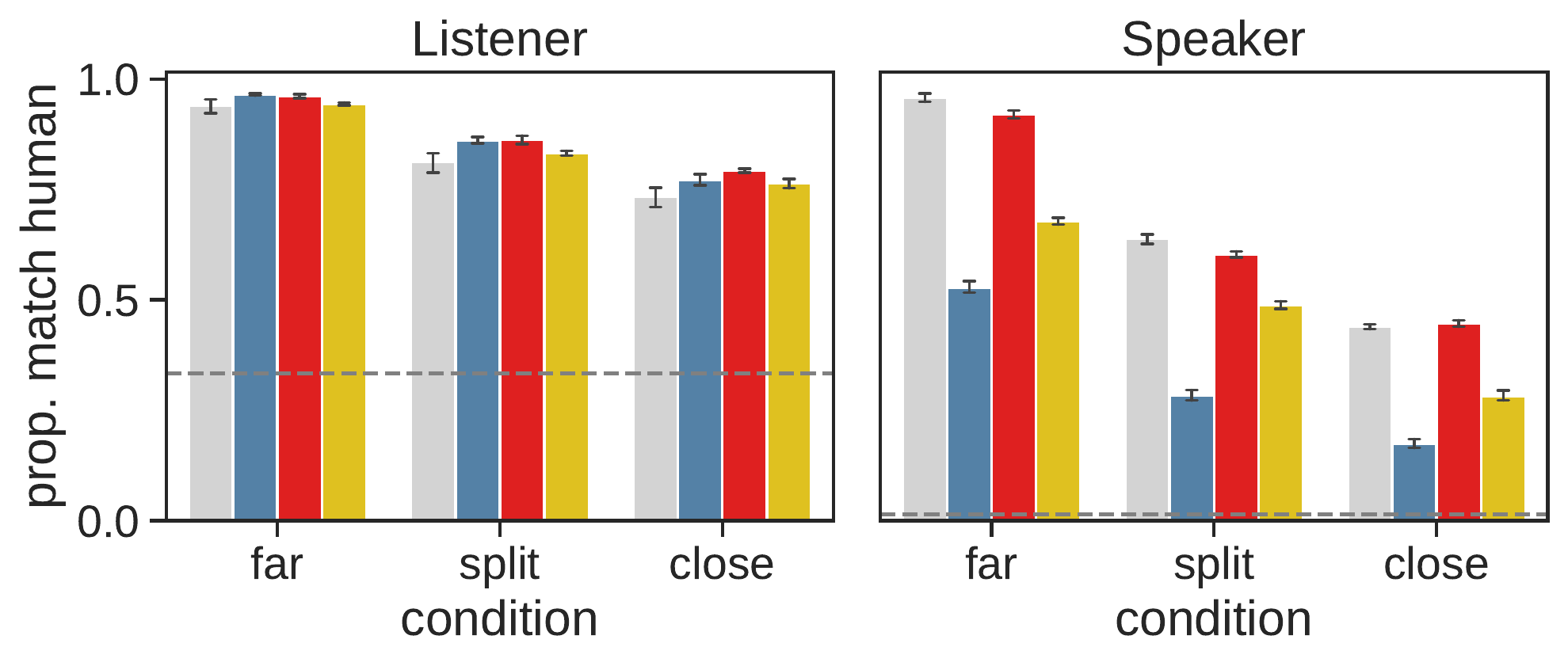}\label{fig:fit}}\vspace{-1em}}\qquad
    \caption{(a) Proportion of test set rounds where listener fails to select ground-truth target color. (b) Proportion of test set rounds where model agents match human behaviors. Listeners always receive human speaker utterance.}
    \label{fig:results}
\end{figure*}

\subsection{Supervised learning (SL) baseline}

For comparison, we implement pragmatic listener and speaker models trained with supervised learning (SL) following \citeauthor{mcdowell-goodman-2019-learning}'s (\citeyear{mcdowell-goodman-2019-learning}) approach. To this end, we use the same lexicon architecture and base agents 
described in \Cref{sec:colors-lex,sec:base}, respectively. 
The pragmatic speaker $s_1$ and pragmatic listener $l_1$ are defined as in \Cref{eq:am-rsa-st,eq:am-rsa-lt}, respectively.
While our lexicon in \Cref{sec:colors-lex} was trained without contexts or pragmatic supervision, the supervised lexicon is trained to maximize the likelihood of the \emph{contextualized} dataset under the $l_1$ pragmatic listener. 
We use the same $\alpha$ as in our SSL agents, but tune the learning rate, batch size, and number of epochs. 

The key difference between the SL and SSL agents is that our SSL agents find pragmatic policies without any context-sensitive data -- that is, without observing human pragmatic behavior. In addition, our SSL agents only need human data to learn literal semantics,
whereas the SL agents fit the ``full stack'' of semantics and pragmatics to human behavior. 

\section{Results}
\Cref{fig:l-acc} shows the proportion of game rounds in the held-out test set where listeners failed to choose the ground-truth target color, given the human speaker utterance. In line with prior work, humans make few mistakes, and error rates increase across the board as the task becomes more difficult (\emph{far} $\to$ \emph{close}). Taking accuracy to be $1 - $ error rate, the pragmatic models achieve accuracy comparable to that reported by \citet{mcdowell-goodman-2019-learning} (87.71\%) and \citet{monroe-etal-2017-colors} (86.98\%),\footnote{Our results are not strictly comparable to theirs, as we use a simpler dataset and different agent architectures.} and yield numerical improvements upon the base model (base 83.4\%; SSL-AM 87.3\%; SSL-GD 87.4\%; SL 85.7\%). 
Within the pragmatic models, we find a significant difference between SL accuracy and the mean accuracy of the SSL models ($t=-4.720,p<10^{-6}$), demonstrating that self-supervised models can outperform models trained to imitate pragmatic human behavior. We find no significant difference between SSL-AM and SSL-GD ($t=0.115,p=0.9$) in our controlled dataset.\footnote{We included model class as a fixed effect, with a random intercept for seed and random slopes for context and game. $p$-values were computed with \texttt{lmerTest} \citep{kuznetsova_lmertest_2017}.}

\Cref{fig:fit} shows the percentage of game rounds where models assigned highest probability to the color/utterance chosen by the human listener/speaker. Since the listener goodness-of-fit results are similar to those of \Cref{fig:l-acc}, we focus on the speaker's fit (right panel). Among the pragmatic agents, the SSL-GD speaker achieves the best fit to the human speaker.
Surprisingly, however, although the pragmatic listener agents improve communicative success over the base listener (\Cref{fig:l-acc}), the pragmatic speakers lead to a significant decrease in fit compared to the base speaker ($p<10^{-6}$; SSL-GD $t=-4.72$, SSL-AM $t=-26.14$, SL $t=-21.52$).
We speculate that SSL-AM may be exploiting the gradedness of the neural lexicon, resulting in pragmatic drift \citep{lazaridou_multi-agent_2020}. In the examples shown in \Cref{tab:contexts}, the SSL-AM speaker chooses an utterance that may not be semantically conventional, but is pragmatically prudent: while ``gray'' sufficiently distinguishes the target color $m^*$ in $C_1$, ``blue'' increases the margin from the other colors (and analogously with ``red'' in $C_2$). 
Due to the sparseness of the dataset, it is unknown how human listeners would interpret model utterances in cases where they differ from human utterances. We leave this for future research. 

\section{Discussion} \label{sec:discussion}

We have proposed a scalable self-supervised approach to pragmatic reasoning, in which agents learn pragmatic policies by optimizing an intrinsic objective instead of imitating human behavior. Our SSL-GD model is comparable to 
supervised learning models for a color reference task. It achieves this without contextualized training data, and  employs a scalable architecture that cannot be used in models based on the traditional AM RSA approach. This work suggests several future research directions that we intend to explore. First, our SSL-GD approach can help study how 
pragmatic knowledge might be shared across contexts and domains, by allowing the agents' parameters to continuously adapt.
Second, the fact that our models do not need contextualized human data expands the range of datasets that can be used for training.
Finally, we would like to test more complex domains and architectures in the interest of ecological validity.
More broadly, our models can be seen as executing a form of algorithmic computation \citep{velickovic_neural_2021} that is grounded in pragmatic theory and information theory, motivating further research into the principles that give rise to human pragmatic reasoning.

\section*{Acknowledgements}



JH was supported by an NSF Graduate Research Fellowship. RPL acknowledges support from NSF grants BCS-1551866 and BCS-1456081, a Google Faculty Research Award, Elemental Cognition, and the MIT Quest for Intelligence. NZ was supported by a BCS Fellowship in Computation. 


\bibliography{example_paper}
\bibliographystyle{icml2021}

\end{document}


\begin{center}
    \Large \textbf{Supplementary material for}\\\textbf{``Scalable pragmatic communication via self-supervision''}
\end{center}

\section*{Gradients} \label{sec:gradients}
\newcommand{\Fa}{\mathcal{F}_{\alpha}}
\newcommand{\Ga}{\mathcal{G}_{\alpha}}
\newcommand{\M}{\mathcal{M}}
\newcommand{\U}{\mathcal{U}}
\newcommand{\Spar}{S_{\psi}}
\newcommand{\Lpar}{L_{\varphi}}
\newcommand{\dphi}{\frac{\partial}{\partial\varphi}}
\newcommand{\dpsi}{\frac{\partial}{\partial\psi}}

For a particular context $C$ with meanings $\M$ and utterances $\U$, the LE-RSA objective with respect to a parameterized listener $\Lpar$ and speaker $\Spar$ is given by
\begin{align}
    \mathcal{G}_{\alpha}(\varphi,\psi) &= H_{\Spar}(U|M) + \alpha \mathbb{E}_{\Spar}\left[\log \Lpar(M|U)\right] \\
    &= -\sum_{m\in\M} P(m) \sum_{u\in\U} \Spar(u|m) \left(\log \Spar(u|m) - \alpha \log \Lpar(m|u)\right) \label{eq:rsa-exact}
\end{align}
and the RD-RSA objective is given by
\begin{align}
    \mathcal{F}_{\alpha}(\varphi,\psi) &= I_{\Spar}(M;U) - \alpha \mathbb{E}_{\Spar}\left[\log \Lpar(M|U)\right] \\
    &= \sum_{m\in\M} P(m) \sum_{u\in\U} \Spar(u|m) \left(\log \frac{\Spar(u|m)}{\Spar(u)} - \alpha \log \Lpar(m|u)\right) \label{eq:rdrsa-exact}.
\end{align}
\paragraph{LE-RSA.} First, we derive the gradients with respect to both sets of parameters for LE-RSA. Note that the LE-RSA objective should be \emph{maximized}, so the gradients should be negated for GD.
\begin{align}
    \frac{\partial \Ga}{\partial \varphi} &= - \sum_{m,u} P(m) \Spar(u|m)\dphi \left[\log \Spar(u|m) - \alpha \log\Lpar(m|u)\right] \\
    &= \alpha \sum_{m,u} P(m) \Spar(u|m) \frac{\partial}{\partial\varphi}\log\Lpar(m|u) \label{eq:grad-rsa-l} \\ 
    \frac{\partial \Ga}{\partial \psi} &= -\sum_{m,u} P(m) \left(\Spar(u|m)\dpsi\left[\log\Spar(u|m)-\alpha\log\Lpar(m|u)\right] + \left(\log\Spar(u|m)-\alpha\log\Lpar(m|u)\right)\dpsi\Spar(u|m)\right) \\
    &= -\sum_{m,u} P(m) \left(\Spar(u|m)\dpsi\log\Spar(u|m) + \left(\log\Spar(u|m)-\alpha\log\Lpar(m|u)\right)\dpsi\Spar(u|m)\right) \\
    &= -\sum_{m,u} P(m) \left(\dpsi\Spar(u|m) + \left(\log\Spar(u|m)-\alpha\log\Lpar(m|u)\right)\dpsi\Spar(u|m)\right) \\
    &= -\sum_{m,u} P(m) \left(\log\Spar(u|m)-\alpha\log\Lpar(m|u)+1\right)\dpsi\Spar(u|m) \label{eq:grad-rsa-s}
\end{align}
\paragraph{RD-RSA.} The gradient for the RD-RSA listener is the inverse of the gradient for the LE-RSA listener (\Cref{eq:grad-rsa-l}), so we turn to the speaker.
\begin{align}
    \frac{\partial \Fa}{\partial \psi} &= \sum_{m,u} P(m) \left(\Spar(u|m)\dpsi\left[\log\frac{\Spar(u|m)}{\Spar(u)}-\alpha\log\Lpar(m|u)\right] + \left(\log\frac{\Spar(u|m)}{\Spar(u)}-\alpha\log\Lpar(m|u)\right)\dpsi\Spar(u|m)\right) \\
    &= \sum_{m,u} P(m) \left(\Spar(u|m)\dpsi\log\frac{\Spar(u|m)}{\Spar(u)} + \left(\log\frac{\Spar(u|m)}{\Spar(u)}-\alpha\log\Lpar(m|u)\right)\dpsi\Spar(u|m)\right) \\
    &= \sum_{m,u} P(m) \left(\dpsi\Spar(u|m)-\Spar(u|m)\dpsi\log\Spar(u) + \left(\log\frac{\Spar(u|m)}{\Spar(u)}-\alpha\log\Lpar(m|u)\right)\dpsi\Spar(u|m)\right) \\
    &= \sum_{m,u} P(m) \left(\left(\log\frac{\Spar(u|m)}{\Spar(u)}-\alpha\log\Lpar(m|u)+1\right)\dpsi\Spar(u|m)-\Spar(u|m)\dpsi\log\Spar(u)\right)
\end{align}